\documentclass{article} 
\PassOptionsToPackage{table}{xcolor}
\usepackage{iclr2027_conference,times}

\usepackage{amsmath,amsfonts,bm}

\def\eqref#1{equation~\ref{#1}}

\def\1{\bm{1}}

\DeclareMathAlphabet{\mathsfit}{\encodingdefault}{\sfdefault}{m}{sl}
\SetMathAlphabet{\mathsfit}{bold}{\encodingdefault}{\sfdefault}{bx}{n}

\renewcommand{\eqref}[1]{(\ref{#1})}

\usepackage{amsmath}
\usepackage{graphicx}
\usepackage{xcolor}
\definecolor{oursblue}{RGB}{218,237,244}
\usepackage{booktabs}
\usepackage{multirow}
\usepackage{wrapfig}
\usepackage{float}
\usepackage{enumitem}
\usepackage{tcolorbox}
\usepackage{caption}
\usepackage{url}
\usepackage{hyperref} 

\newtcolorbox{questionbox}[1][]{
    colback=blue!3!white,
    colframe=black,
    fonttitle=\bfseries\small,
    title={#1},
    boxrule=0.5pt,
    left=6pt, right=6pt, top=4pt, bottom=4pt
}

\newtcolorbox{responsebox}[1][]{
    colback=white,
    colframe=black!30,
    boxrule=0.4pt,
    left=6pt, right=6pt, top=4pt, bottom=4pt
}

\title{Plug-and-Play Verifier for Decoupling Process Reward in Stabilizing Long-Horizon Agentic Reasoning}

\author{
\textbf{Mingwei Xu}\textsuperscript{1},
\textbf{Jieyuan Pei}\textsuperscript{2},
\textbf{Binghao Tu}\textsuperscript{3},
\textbf{Hao Fang}\textsuperscript{1,*}\\[0.5em]
\textsuperscript{1}Department of Electrical and Computer Engineering,
University of Washington, Seattle, WA, USA\\
\textsuperscript{2}The Hong Kong University of Science and Technology (Guangzhou),
Guangzhou, China\\
\textsuperscript{3}Department of Electrical Engineering,
City University of Hong Kong, Hong Kong SAR, China\\
\textsuperscript{*}Corresponding author\\
\texttt{\{xmw03, haof459\}@uw.edu}
}

\iclrfinalcopy 
\begin{document}
\maketitle

\begin{center}
  \captionsetup{type=figure}
  \includegraphics[width=1.0\linewidth]{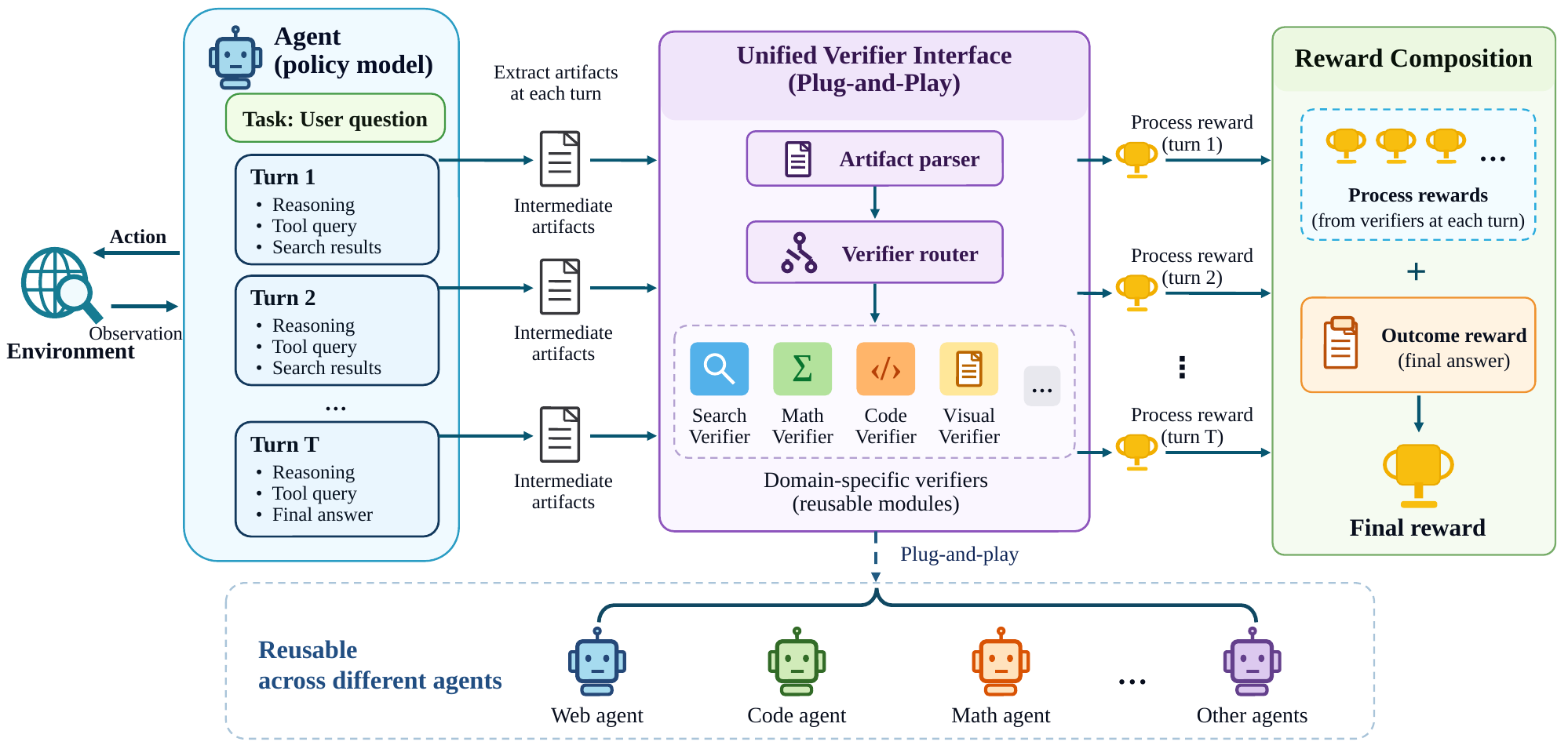}
  \caption{\textbf{Overview of the plug-and-play verifier.} \textit{Left:} artifacts are extracted per interaction step. \textit{Middle:} a unified interface routes them to domain-specific verifiers, which return verification scores assigned to source steps. \textit{Right:} These final reward are aggregated into process rewards and combined with the outcome reward. \textit{Bottom:} verifiers interact with RL post-training through the interface and can be used independently across agents.}
  \label{fig:teaser}
\end{center}

\begin{abstract}
Stabilizing long-horizon reasoning requires intelligent agents to sustain useful progress across successive reasoning steps and tool interactions. Such agents are typically trained with reinforcement learning (RL), where actor-critic RL uses a learnable critic network to estimate step advantages, while critic-free methods such as GRPO eliminate critic models and rely mostly on the group variants of sparse outcome-only rewards. Both reward types leave partially useful failed trajectories either implicit or indistinguishable from entirely unproductive ones, and as reasoning horizons grow, it becomes difficult to maintain consistent reasoning trajectories. In this work, we propose a \textit{plug-and-play verifier} framework for stabilizing long-horizon reasoning through explicit verification of intermediate reasoning artifacts. Specifically, the framework decouples process verification through a unified interface that extracts intermediate reasoning artifacts from agent running trajectories, evaluates them using domain-specific verifiers, and aggregates the verification scores as bounded process reward. Then, we combine our dense verified process rewards and the sparse outcome rewards through a two-stage post-training schedule. We showed that with plug-and-play verifiers, we further stabilize the reasoning steps as last-mile self-reflection and self-consistency. Typical evaluation on HotpotQA using Qwen3.5-4B achieves a superior accuracy (60.18\%), outperforming the standard GRPO baseline (57.73\%). Our results suggest that plug-and-play verifiers provide a modular approach to learn more stable and reliable reasoning behaviors over long horizons.\footnote{\url{https://github.com/momo1443/plug-and-play-verification}}
\end{abstract}

\section{Introduction}
\label{sec:introduction}

Large language models are increasingly deployed as agents that solve tasks by interleaving
reasoning with actions in external environments, following the reasoning, acting, and observing
cycle proposed by ReAct~\citep{yao2022react}. Equipped with tools and environment feedback, these agents tackle information seeking, software engineering, and web navigation through sequences of interdependent decisions~\citep{schick2023toolformer,yang2024swe,zhou2024webarena,zheng2025deepresearcher}.
Success on these tasks requires effective agentic reasoning, where agents must interpret newly acquired evidence, revise their plans, and carry stable intermediate results forward to guide subsequent actions~\citep{shinn2023reflexion,zhou2023language,zhu2026dense}. However, sustaining this capability becomes increasingly challenging as the interaction horizon grows~\citep{da2025agent,gandhi2026draco}.
For example, a hallucinated claim, calculation error, or misinterpreted tool result can misdirect subsequent actions as later decisions build on earlier reasoning and observations~\citep{yao2022react,yuan2026verifiable}.
Evaluations across interactive environments, software engineering, and general assistant tasks further highlight persistent difficulties in reliably completing multi-step tasks~\citep{liu2024agentbench,jimenez2024swe}. These difficulties are exacerbated in multi-agent interactions, where an agent's erroneous outputs are passed to others and incorporated into reasoning, allowing errors to propagate and compound across agents~\citep{du2023improving,cui2026free}. Therefore, stabilizing long-horizon agentic reasoning is emergent~\citep{xu2026stamp}.

Reinforcement learning provides an effective approach to improving agentic reasoning by allowing models to explore alternative strategies and learn from verifiable outcomes without requiring human annotations for every intermediate step~\citep{wang2025ragen,guo2025deepseek,jin2025search,feng2026retool}. Both actor-critic methods and critic-free methods such as GRPO rely on the supplied rewards to guide policy optimization~\citep{schulman2017proximal,shao2024deepseekmath}.
However, outcome-only rewards provide limited information about intermediate progress, assigning identical feedback to trajectories that may differ substantially in the validity of their intermediate artifacts~\citep{yang2026progress,zhang2025lessons,xu2026beyond}. This motivates explicit process supervision that can distinguish the quality of intermediate artifacts in long-horizon agentic tasks~\citep{yang2026progress,zhang2025lessons,yuan2026verifiable}.

In this work, we observe that process rewards can be obtained by checking intermediate artifacts against task conditions and evidence independently of reward composition and policy optimization. Specifically, assessing whether an artifact satisfies a verification criterion does not require deciding how the score is combined with outcome rewards or used to update the policy. Conversely, downstream training requires the verification scores and their source-step assignment from the verifier. This independence allows us to decouple process verification from reward composition and policy optimization. Therefore, we implement process verification in external modules that interact with training through an interface that passes verification scores and their source-step assignment. Moreover, agents in different domains produce different intermediate artifacts. To support agents without adapting the training procedure to each verifier, we propose a unified plug-and-play verification interface. As illustrated in Figure~\ref{fig:teaser}, every verifier module extracts intermediate artifacts, verifies them, and returns bounded weighted sums of verification scores assigned to source interaction steps. Since downstream reward composition and policy optimization depend on each verifier only through these outputs, verifier modules for different agents can be introduced or replaced in a plug-and-play manner. Inspired by the above, we propose a \textit{plug-and-play verifier} framework for stabilizing long-horizon agentic reasoning through explicit verification of intermediate artifacts. Our framework makes the following contributions:
\begin{itemize}[topsep=2pt,itemsep=2pt,parsep=0pt]

\item Process rewards by verifying intermediate artifacts against task conditions and environment evidence, providing complementary supervision beyond outcome rewards.

\item A unified interface that decouples process verification from reward composition and policy optimization, enabling plug-and-play integration of verifiers.]
\end{itemize}

To evaluate it, we conduct experiments using Qwen3.5-4B and Qwen3.5-9B on the HotpotQA validation split, LiveCodeBench, MATH-Vision, and AIME 2026, covering multi-hop question answering, code generation, multimodal reasoning, and mathematical reasoning. Our experiments show that incorporating a plug-and-play verifier can improve the performance and stabilize the reasoning trajectories. Our ablation studies further examine verification mechanisms and support the contributions of reward composition. Our results suggest that explicit process verification can complement outcome rewards to improve agentic reasoning.
\vspace{-1mm}

\section{Related Work}
\label{sec:relatedwork}
\textbf{Long-horizon reasoning and multi-agent interaction.}
Long-horizon agentic reasoning requires agents to make successive decisions based on previously generated information and observations from tools or environments. Early approaches support multi-step reasoning through explicit intermediate steps and exploration of alternative reasoning paths~\citep{wei2022chain,yao2023tree}. Tool-augmented approaches connect reasoning with external information and executable actions, enabling agents to interact with environments and perform tasks~\citep{yao2022react,schick2023toolformer,yang2024swe}. Feedback and episodic memory can guide agents to revise their strategies across successive attempts~\citep{shinn2023reflexion}. Multi-agent approaches extend these capabilities through role-specific workflows and iterative debate~\citep{hong2024metagpt,du2023improving}. However, evaluations on WebArena and AgentBench reveal persistent difficulties in long-term planning, decision-making, and reliable task completion~\citep{zhou2024webarena,liu2024agentbench}. Analyses of multi-agent systems identify failures in coordination and task verification~\citep{cemri2026multi}. Recent analysis of search agents further identifies failures in evidence propagation~\citep{sang2026searchatlas}.

\textbf{Agentic reinforcement learning.}
Reinforcement learning enables agents to learn reasoning and tool-use strategies from task feedback. Search-R1 and ReTool incorporate search and code execution into multi-turn trajectories and optimize interaction policies using outcome rewards~\citep{jin2025search,feng2026retool}. Actor-critic methods use learned value functions to estimate advantages; these functions predict returns rather than verify intermediate artifacts~\citep{schulman2017proximal}. Critic-free methods such as GRPO normalize rewards within a group of trajectories, but under outcome-only supervision, each trajectory receives a single outcome reward~\citep{shao2024deepseekmath}. Research on multi-turn agent training addresses training stability and temporal credit assignment~\citep{wang2026arlarena,wang2026ragen,he2026hierarchy,yang2026progress}. More recently, GiGPO combines trajectory-level comparisons with step-level advantage estimation over actions taken at recurring environment states~\citep{feng2026group}. IGPO instead derives step-level rewards from increases in the policy's probability of producing the correct answer and combines them with outcome supervision~\citep{wang2025information}.

\textbf{Process supervision and verifiable feedback.}
Process supervision provides dense feedback on intermediate reasoning beyond sparse final outcome correctness. A central question is how to obtain reliable supervision for these intermediate steps. Initial human-in-the-loop methods demonstrate the effectiveness of the supervision in mathematical reasoning~\citep{lightman2024let}, while Math-Shepherd and ReST-MCTS derive supervision automatically from sampled trajectories and tree search~\citep{wang2024math,zhang2024rest}. Rewarding Progress further evaluates intermediate steps by their contribution to a prover policy's likelihood of success~\citep{setlur2025rewarding}. Alongside these approaches, another line of work grounds feedback in tools, evidence, or explicit verification criteria. CRITIC uses tools to validate and revise outputs at inference time~\citep{gou2024critic}, while VPR and VPRMs incorporate symbolic or algorithmic checks and deterministic rules into process rewards for intermediate actions and structured reasoning trajectories~\citep{yuan2026verifiable,pronesti2026beyond}. For agentic tasks, this raises a further question of how to attribute feedback to individual actions~\citep{gandhi2026draco,chai2025rlfactory}. STAMP attributes verified evidence to the actions that first exposed it, while ABSeeker constructs step-level supervision from answer-backtracked clues~\citep{xu2026stamp,lu2026abseeker}.

\section{Methods}
\label{sec:method}

\subsection{Problem Setting and Preliminaries}
\label{subsec:problem_setting}

We study long-horizon agentic reasoning tasks.
Given a task domain $d$, let $x\sim\mathcal{D}_d$ denote an input consisting of a task instruction and any available context, including images for multimodal tasks.
A policy $\pi_\theta$, parameterized by $\theta$, produces a solution through successive policy generations, and interactions with an external environment.

\paragraph{Policy and trajectories.}
At interaction step $t$, the policy generates an action $a_t\sim\pi_\theta(\cdot\mid h_t)$ conditioned on the interaction history $h_t=(x,a_1,o_1,\ldots,a_{t-1},o_{t-1})$.
Specificall, an action is a complete model-generated response that can contain reasoning text, a tool call, or a final answer.
When a tool is invoked, the environment returns an observation $o_t$, which is incorporated into the history for the next step.
The interaction proceeds until the agent submits a final answer or reaches a stopping condition, producing a trajectory
\begin{equation}
\tau=(a_1,o_1,\ldots,a_{T-1},o_{T-1},a_T),
\label{eq:agent_trajectory}
\end{equation}
where $T$ denotes the number of policy generations.
Observations can include tool outputs or continuation prompts supplied by the interaction protocol, and are omitted when no such input is supplied.
Each policy generation contain multiple artifacts.
Verification is performed at the artifact level, with scores attributed to the artifacts' source interaction steps.

\paragraph{Outcome and process rewards.}
Trajectories with the same outcome reward may differ in whether their intermediate reasoning artifacts satisfy explicit task conditions and are supported by recorded evidence.
Thus, we distinguish outcome and process rewards:
\begin{equation}
E_d(x,\tau)\in[0,1],
\qquad
P_d(x,\tau)\in[0,1].
\label{eq:reward_components}
\end{equation}
The outcome reward $E_d(x,\tau)$ evaluates the final answer against domain-specific correctness, and is zero if no valid final answer is submitted.
Evaluation references are available to the evaluator and need not be included in the policy input.
In contrast, the process reward $P_d(x,\tau)$ instead aggregates verification scores for intermediate artifacts, as defined in Section~\ref{subsec:process_verifiers}.
Importantly, a passing local check establishes only the property tested by its rule and does not establish final-answer correctness.

\subsection{Plug-and-Play Process Verification}
\label{subsec:process_verifiers}
Our framework decouples domain-specific artifact verification from reward composition and policy optimization through a unified interface.

\paragraph{Decoupled local verification.}
Verification assesses an artifact from explicitly specified inputs, without evaluating the full reasoning narrative.
To this end, a domain-specific extractor identifies $m$ artifacts in $(x,\tau)$ and their source interaction steps.
An artifact is an explicit object admitting a check, such as an arithmetic equality, a citation, or a code version associated with an execution record.
For each extracted artifact $c_j$, the verification score is computed as
\begin{equation}
v_j=V_d(c_j;\xi_j)\in[0,1],
\label{eq:local_verification}
\end{equation}
where $\xi_j$ contains the task conditions and records required by the check.
These can include earlier artifacts or specified submission fields.
For fixed rules and evaluation conditions, the score depends on the trajectory only through $(c_j,\xi_j)$.
Consequently, changing unrelated reasoning text or the final correctness label does not change the score.
Moreover, the verifier does not access policy parameters or hidden states, sample policy continuations, or estimate eventual task success.

This separation allows checks to be replayed after trajectory completion when their required inputs have been recorded.
In our implementation, the QA verifier uses source-span grounding and action-certificate matching without annotated supporting facts; the code verifier uses version lineage and developer-test replay; the visual verifier checks crop replay and string-level answer-claim coupling; and the mathematics verifier checks extracted numeric equalities.
These fixed rules verify their stated properties rather than all reasoning steps.
Representative verification inputs and checks for the four domains are provided in Appendix~\ref{app:example_responses}.

\paragraph{Plug-and-play integration.}
We expose local verification through adapters and a common verification interface.
Specifically, Let $B_\ell$ be a group of checks assigned to source step $s_\ell$, with disjoint groups and domain-specific nonnegative weights.
For each group, $q_\ell$ is the weighted sum of verification scores assigned to source step $s_\ell$:
\begin{equation}
q_\ell=\sum_{j\in B_\ell}\omega_{d,j}(x,\tau)v_j,
\qquad
\omega_{d,j}\geq0,
\qquad
\sum_{j=1}^{m}\omega_{d,j}\leq1.
\label{eq:local_score_to_credit}
\end{equation}
Here, the weights encode domain-specific normalization and check applicability.
In particular, interaction steps without applicable verification checks are excluded from step-level normalization, whereas applicable checks that fail verification contribute zero scores.
Artifacts scored individually use singleton groups.
Unlike the local verification score $v_j$, the weights used to compute $q_\ell$ can depend on other artifacts in the trajectory.
Thus, the common bound provides a shared reward range, while the verification criteria and normalization rules remain domain-specific.

\begin{figure}[!htbp]
    \centering
    \includegraphics[width=1.0\textwidth]{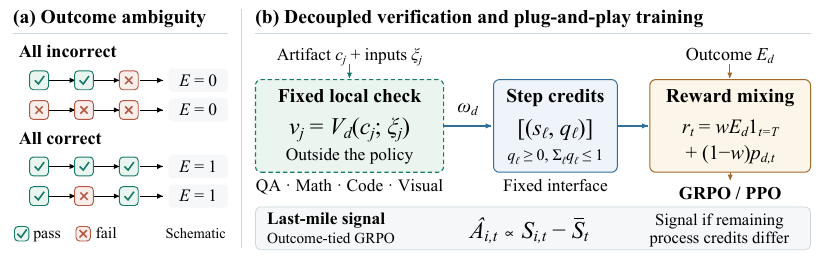}
    \caption{The plug-and-play process verification.
(a) Identical outcome rewards can mask differences in
verification results.
(b) Local checks are decoupled from policy reasoning and return
weighted sums of verification scores assigned to source steps
through a common interface.
These are aggregated into process rewards and combined with
outcome rewards for GRPO or PPO training.}
    \label{fig:framework}
\end{figure}

Based on these aggregated scores, the complete verifier module returns a set of source-step assignments and verification scores:
\begin{equation}
\Phi_d(x,\tau)=\bigl[(s_\ell,q_\ell)\bigr]_{\ell=1}^{n},
\qquad
s_\ell\in\{1,\ldots,T\},
\quad
q_\ell\geq0,
\quad
\sum_{\ell=1}^{n}q_\ell\leq1.
\label{eq:verifier_output_contract}
\end{equation}
Here, $n$ is the number of emitted records.
Records with $q_\ell=0$ may be omitted, while full audits retain failed checks, including applicable checks with missing required evidence.
An empty verifier output yields zero process reward.
Importantly, downstream reward composition consumes only $(s_\ell,q_\ell)$, allowing compatible verifier modules to be replaced without modifying reward composition or policy optimization.
Reuse with a different agent requires compatible artifacts and an adapter for its records.
Figure~\ref{fig:framework} summarizes this decoupled design.

\paragraph{Reward composition and optimizer interface.}
\label{subsec:reward_composition}
At interaction step $t$, the process reward is $p_{d,t}(x,\tau)=\sum_{\ell:s_\ell=t}q_\ell$, with empty sums equal to zero.
To combine process and outcome supervision, we introduce an outcome-process weight coefficient $w \in [0,1]$ and define the step reward as
\begin{equation}
r_t =
w\mathbf{1}\{t=T\}E_d(x,\tau)
+(1-w)p_{d,t}(x,\tau).
\label{eq:reward_composition}
\end{equation}
Process rewards are retained for independently verifiable intermediate artifacts whether a valid final answer is submitted.
The outcome reward is assigned to the terminal step, while process rewards are assigned to their source steps.
With $P_d(x,\tau)=\sum_t p_{d,t}(x,\tau)$, the trajectory-level reward is $R_{d,w}(x,\tau)=\sum_t r_t=wE_d(x,\tau)+(1-w)P_d(x,\tau)$.
The final rewards can be used by GRPO or PPO without changing the verification rules.
In practice, the training uses outcome-only rewards ($w=1$) for the first 50 updates.
At each subsequent update, a weight is sampled reproducibly from $\mathcal U(0,1)$ for each prompt group by all rollouts in that group.
More details are provided in Appendix~\ref{app:policy_optimization}.

\subsection{Last-Mile Refinement}
\label{subsec:last_mile}
Trajectories with identical outcome rewards can differ in the validity of their intermediate reasoning artifacts.
This difference allows process rewards to provide last-mile refinement signal when sparse outcome-only supervision is identical~\citep{li2026no}.

\paragraph{Differentiate rewards for policy refinement.}
Outcome-only GRPO relies on reward differences within a group of rollouts to construct relative advantages.
With binary outcome rewards, both all-incorrect and all-correct groups have zero reward variance~\citep{yu2026dapo}.
All-incorrect groups can impede early learning when the initial policy rarely generates successful rollouts.
Cold-start supervised fine-tuning (SFT) on reasoning demonstrations provides an initialization strategy for RL~\citep{guo2025deepseek}. On the other hand, the same issue also happens in correct groups.
For a fixed prompt $x$ and interaction step $t$, we consider a group $\mathcal G_t$ with outcome rewards $E$ and a shared weighted coefficient $0\leq w<1$.
All rollouts in $\mathcal G_t$ receive identical outcome rewards yielding zero advantages.
Let $S_{i,t}=\sum_{u=t}^{T_i}p_{d,u}(x,\tau_i)$ denote the remaining process return, and let $\mathbf S_t=(S_{i,t})_{i\in\mathcal G_t}$.
Using the reward composition in Eq.~\eqref{eq:reward_composition}, the weighted return is $R_{i,t}=wE+(1-w)S_{i,t}$.
Subtracting the group of rollout mean removes the common outcome term, yielding
\begin{equation}
\widehat A_{i,t}=
\frac{S_{i,t}-\operatorname{mean}(\mathbf S_t)}
{\operatorname{std}(\mathbf S_t)
+\epsilon/(1-w)},
\label{eq:tied_outcome_advantage}
\end{equation}
where $\epsilon>0$ ensures the numerical stability and $\operatorname{std}$ is the sample standard deviation.
Thus, differences in remaining process returns yield nonzero group-relative advantages even if rewards on all rollouts are the same.
These differences provide the \emph{last-mile refinement signal} for refining intermediate artifacts when the group's final answers are all correct.
More details are in Appendix~\ref{app:policy_optimization}

\paragraph{Verification-privileged self-consistency.}
The last-mile refinement signal is grounded in external verification of intermediate artifacts.
The verifiers evaluate these artifacts against explicit recorded evidence.
Therefore, the policy learns from process rewards derived from these checks rather than judgments from a black-box reward model. Here, we interpret self-consistency from the above external verification perspective.
Specifically, verification-privileged self-consistency requires all applicable verification checks within a single trajectory to pass.
Let $\mathcal J_d(x,\tau)$ index the full set of applicable checks under a fixed evaluation protocol.
A check fails if its required evidence is missing.
Let $\mathcal J_{d,+}(x,\tau)= \{j\in\mathcal J_d(x,\tau)\mid v_j=1\}$ denote the set of indices of checks passed by the trajectory.
We define
\begin{equation}
C_d(x,\tau)=
\mathbf{1}\!\left\{
\bigl(\mathcal J_d(x,\tau)\neq\emptyset\bigr)
\land
\bigl(\mathcal J_{d,+}(x,\tau)=\mathcal J_d(x,\tau)\bigr)
\right\}.
\label{eq:verified_self_consistency}
\end{equation}
Thus, $C_d(x,\tau)=1$ if and only if at least one check applies and every applicable check has verification score $v_j=1$.
This criterion essentially measures internal (within-trajectory) consistency that obviously differs from the most recent majority-vote-based (external) work via selecting the most frequent answers in sampled trajectories~\citep{wang2022self}.

\paragraph{Verifier-induced self-reflection.}
Moreover, internal self-consistency motivates self-reflection, where we interpret self-reflection as the policy proactively checking and revising its intermediate reasoning artifacts. Intuitively, self-consistency encourages self-reflection by consistently reducing generated hallucinations at intermediate reasoning steps.
After trajectory generation, the verifier checks recorded intermediate artifacts and assigns process rewards to their source steps.
An earlier failed check remains in $\mathcal J_d(x,\tau)$ and keeps $C_d(x,\tau)=0$, even if the agent later corrects the error.
Nevertheless, these later artifacts can still receive positive process rewards when they passes verification.
Therefore, this step-level supervision can encourage the policy to check and revise intermediate reasoning in future trajectories, potentially reducing error propagation.

Overall, our framework decouples the verification process via independently checking its intermediate reasoning artifacts. When reasoning processes have heterogeneous returns, our approach provides a last-mile supervision signal for refining intermediate artifacts. Futher, the resulting training outcomes ensure self-consistency and self-reflection. Thus, our framework can mitigate error propagation across successive reasoning steps and tool interactions, ultimately supporting more stable long-horizon agentic reasoning (see Appendix~\ref{app:example_responses} for reasoning examples).

\section{Results}
\label{sec:results}
\subsection{Experimental Setup}
\label{subsec:setup}
\textbf{Training Data and Evaluation Benchmarks.}
We evaluate our approach across four domains: multi-hop question answering, code generation, multimodal reasoning, and mathematical reasoning.
For multi-hop question answering, we train on the training split of \textbf{HotpotQA}~\citep{yang2018hotpotqa} and evaluate on its validation split.
For code generation, we train on \textbf{TACO}~\citep{li2023taco} and evaluate on \textbf{LiveCodeBench}~\citep{jain2025livecodebench}.
For multimodal reasoning, we use the \textbf{Vision-R1}~\citep{huang2026vision} for training and \textbf{MATH-Vision}~\citep{wang2024measuring} for evaluation.
For mathematical reasoning, we train on \textbf{DeepScaleR-Preview-Dataset}~\citep{luo2025deepscaler} and evaluate on \textbf{AIME 2026}~\citep{dekoninck2026matharena}.
Training is performed separately for each domain.
Within each domain, all compared methods use the same training data and evaluation split.

\textbf{Baselines.}
We compare our approach with \textbf{ReAct}, which uses the initial model without RL, and \textbf{GRPO}, which trains the same initial model using only the binary outcome reward.
We additionally include \textbf{GRPO + LLM-as-a-judge}, which uses a frozen \textbf{Qwen3.5-9B} judge to provide training rewards for the \textbf{Qwen3.5-4B} policy.
Across the four domains, the judge evaluates intermediate retrieved evidence, code development and testing, visual inspection, and mathematical reasoning using domain-specific criteria.
The resulting process rewards are assigned to the corresponding interaction steps and combined with the same binary outcome reward used by GRPO.
Our method obtains process rewards from domain-specific verifiers instead of an LLM judge.

\textbf{Base Models and Training Framework.}
We use \textbf{Qwen3.5-4B} and \textbf{Qwen3.5-9B}~\citep{qwen3.5} to examine the effectiveness of our approach at two model scales.
Our implementation builds on \textbf{Agent-R1}~\citep{cheng2025agent}, with \textbf{GRPO}~\citep{shao2024deepseekmath} and \textbf{PPO}~\citep{schulman2017proximal} as the training algorithms.
All methods are implemented using the \textit{verl} package for reward computation, training, and generation.
Within each domain and model scale, the RL methods use the same initialization, task interface, tools, fine-tuning method, and training budget.
For all domains, we use the first 10,000 examples from each prepared training split, using global prompt batch sizes of 20 and 8 for Qwen3.5-4B and Qwen3.5-9B.
Qwen3.5-9B uses LoRA with rank $64$ and scaling factor $\alpha=64$.
Domain-specific settings are provided in Appendix~\ref{app:training_details}.

\textbf{Evaluation Protocol.}
We evaluate agents by their final task performance and report accuracy (Acc., \%) across all four benchmarks.
Accuracy is defined as the percentage of evaluation instances for which the agent produces a correct final answer.
Correctness is assessed using benchmark-specific criteria: answer exact match for HotpotQA, passing all evaluation test cases for LiveCodeBench, and answer correctness for MATH-Vision, AIME 2025 and AIME 2026.
For interactive tasks, each evaluation episode includes the complete sequence of agent feedback until termination, and correctness is determined from the final submitted answer.
We use vLLM~\citep{kwon2023efficient} for speed-up generation.

\subsection{Effectiveness Across Domains and Model Scales}
\label{subsec:domains}

\begin{table}
\centering
\renewcommand{\arraystretch}{1.12}
\setlength{\tabcolsep}{5pt}
\resizebox{\textwidth}{!}{%
\begin{tabular}{lcccccc}
\toprule
\textbf{Methods}
& \textbf{AIME 2025}
& \textbf{AIME 2026}
& \textbf{HotpotQA}
& \textbf{LiveCodeBench}
& \textbf{MATH-Vision}
& \textbf{Avg.} \\
\midrule
\multicolumn{7}{@{}l}{\textbf{\textit{Qwen3.5-4B}}} \\
ReAct
& 26.67
& 50.00
& 46.78
& 40.47
& 30.13
& 38.81 \\
GRPO
& 33.33
& 52.50
& 57.73
& 47.96
& 33.22
& 44.95 \\
GRPO + LLM-as-a-judge
& 36.67
& 53.33
& 58.28
& 48.72
& \textbf{35.20}
& 46.44 \\
\rowcolor{oursblue}
\textbf{Ours (plug-and-play verifier)}
& \textbf{40.00}
& \textbf{55.83}
& \textbf{60.18}
& \textbf{49.57}
& 34.54
& \textbf{48.02} \\
\midrule
\multicolumn{7}{@{}l}{\textbf{\textit{Qwen3.5-9B}}} \\
ReAct
& 36.67
& 53.33
& 52.56
& 49.00
& 36.18
& 45.55 \\
GRPO
& 43.33
& 56.67
& 63.20
& 56.59
& 37.01
& 51.36 \\
\rowcolor{oursblue}
\textbf{Ours (plug-and-play verifier)}
& \textbf{46.67}
& \textbf{58.33}
& \textbf{66.67}
& \textbf{58.48}
& \textbf{38.16}
& \textbf{53.66} \\
\bottomrule
\end{tabular}%
}
\caption{
Performance (\%) across four domains using Qwen3.5-4B and Qwen3.5-9B.
Each domain is trained separately on its corresponding training dataset. Bold numbers indicate the best.
}
\label{tab:main_results}
\end{table}

\begin{figure}[ht]
    \centering
    \includegraphics[width=0.95\linewidth]{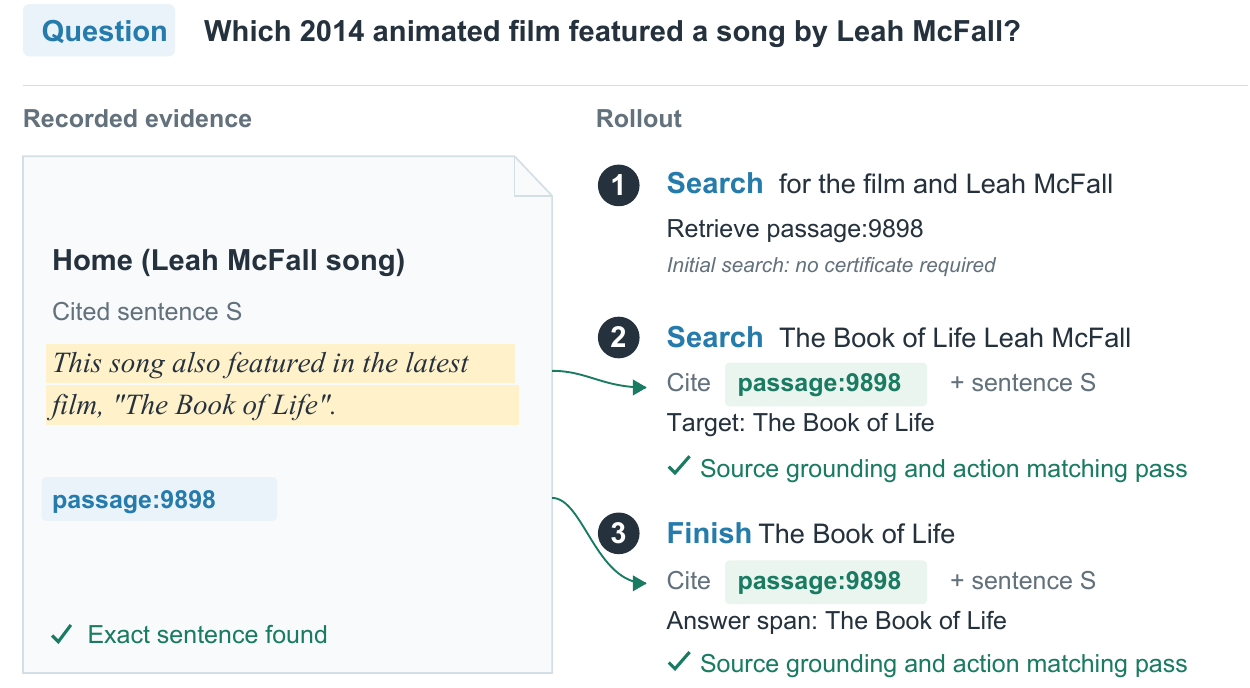}
    \caption{An example of our plug-and-play verifier on HotpotQA.}
    \label{fig:hotpotqa_example}
\end{figure}

We began our experimental analysis with a compact yet effective \textbf{Qwen3.5-4B} model, which has been widely used in recent studies~\citep{yang2026skillopt}.
In Table~\ref{tab:main_results}, we observed that the Qwen3.5-4B model achieved the best performance on four of the five benchmarks (AIME 2025: $40.00$, AIME 2026: $55.83$, HotpotQA: $60.18$, LiveCodeBench: $49.57$) and remained competitive on MATH-Vision ($34.54$).
Furthermore, compared with the baseline methods, our approach achieved the highest average accuracy across the five benchmarks (Ours: $48.02$, vs.\ ReAct: $38.81$, GRPO: $44.95$, GRPO + LLM-as-a-judge: $46.44$).
These results provided initial evidence that our plug-and-play verifier could improve agentic reasoning accuracy by incorporating verified process rewards alongside outcome rewards.

Next, we further tested our plug-and-play verifier using the larger \textbf{Qwen3.5-9B} model (see Table~\ref{tab:main_results}, second block).
First, the large models showed better performance in general.
For example, our method achieved higher accuracy in HotpotQA with Qwen3.5-9B ($66.67$) than with Qwen3.5-4B ($60.18$).
Second, compared with outcome-only GRPO, our method consistently achieved better performance across all five benchmarks (AIME 2025: $46.67$ vs.\ $43.33$, AIME 2026: $58.33$ vs.\ $56.67$, HotpotQA: $66.67$ vs.\ $63.20$, LiveCodeBench: $58.48$ vs.\ $56.59$, MATH-Vision: $38.16$ vs.\ $37.01$).
Meanwhile, our method outperformed both GRPO ($53.66$ vs.\ $51.36$) and ReAct ($53.66$ vs.\ $45.55$) on average.
This trend supported our assumption that sparse outcome rewards may overlook meaningful differences in intermediate artifacts, while explicit process verification can provide complementary learning signals for policy optimization.
We provided an illustrative HotpotQA example in Figure~\ref{fig:hotpotqa_example}, with more details in Appendix~\ref{app:example_responses}, to demonstrate how our plug-and-play verifier evaluates intermediate artifacts.

\begin{figure}[!htbp]
    \centering
    \includegraphics[width=0.95\textwidth]{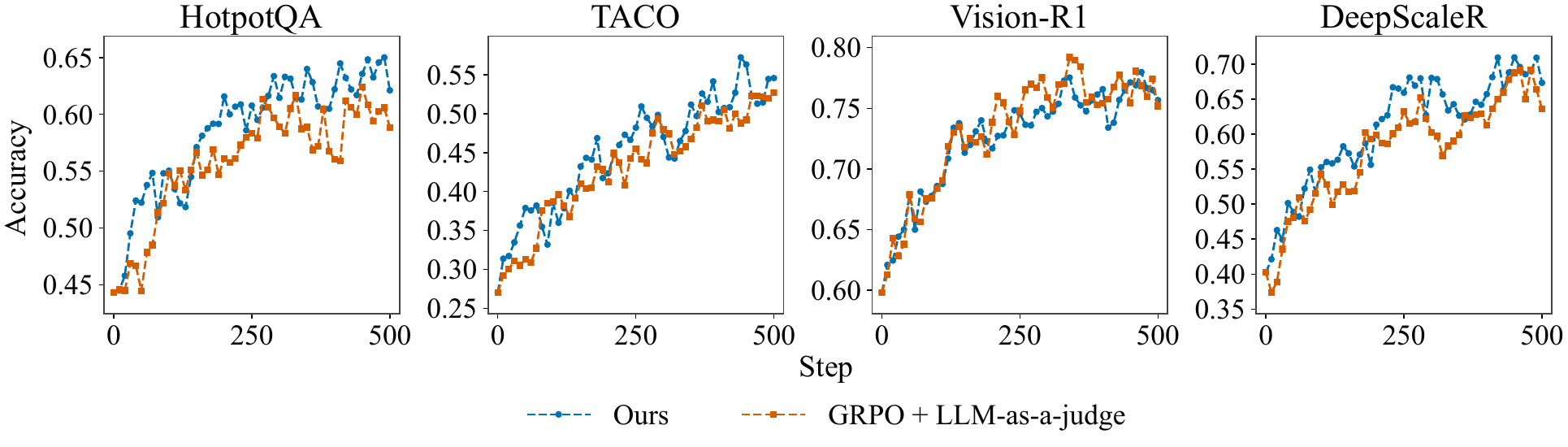}
    \caption{Training accuracy of Qwen3.5-4B with Ours and GRPO with LLM-as-a-judge.}
    \label{fig:training_accuracy}
\end{figure}

Regarding the comparison with the LLM-as-a-judge baseline, while our plug-and-play verifier achieved higher average accuracy using Qwen3.5-4B (Ours: $48.02$ vs.\ LLM-as-a-judge: $46.44$), it slightly underperformed on MATH-Vision ($34.54$ vs.\ $35.20$).
One potential reason was that our visual verifier primarily relied on explicit checks, such as crop provenance, replay consistency, and answer-claim matching, without directly assessing the semantic correctness of visual reasoning.
This might limit the effectiveness of process supervision for tasks requiring more complex visual understanding, where LLM-based verification could provide more flexible feedback.
Figure~\ref{fig:training_accuracy} further compares the training accuracy of both methods across the four training datasets.
One of our future research directions will explore incorporating semantic verification into the proposed plug-and-play framework to further improve its generality across multimodal reasoning tasks.

\subsection{Ablation Studies}
\label{subsec:ablation}
Having established the above results, we examined the interaction between outcome and process rewards and compared different verification mechanisms.
We focused on \textbf{HotpotQA}, which provided both reference answers and annotated supporting facts, allowing outcome supervision and process supervision to be examined separately.
All results in this subsection used \textbf{Qwen3.5-4B} with \textbf{GRPO}.
We reported accuracy based on normalized exact match (EM).

\begin{table}[!htbp]
\centering
\small
\renewcommand{\arraystretch}{1.08}
\begin{minipage}[t]{0.48\linewidth}
\centering
\textbf{(a) Reward configurations}\par
\vspace{0.4em}
\begin{tabular*}{\linewidth}
{@{\extracolsep{\fill}}lr@{}}
\toprule
\textbf{Configuration} & \textbf{Acc. (\%)} \\
\midrule
ReAct & 46.78 \\
Process only & 47.27 \\
Outcome only  & 57.73 \\
Outcome + process & 61.40 \\
\bottomrule
\end{tabular*}
\end{minipage}\hfill%
\begin{minipage}[t]{0.48\linewidth}
\centering
\textbf{(b) Verification mechanisms}\par
\vspace{0.4em}
\begin{tabular*}{\linewidth}
{@{\extracolsep{\fill}}lr@{}}
\toprule
\textbf{Configuration} & \textbf{Acc. (\%)} \\
\midrule
GRPO & 57.73 \\
\quad + LLM-as-a-judge & 58.28 \\
\quad + plug-and-play verification (ours) & 60.18 \\
\quad + Gold evidence (upper bound) & 61.40 \\
\bottomrule
\end{tabular*}
\end{minipage}
\caption{
Ablation studies on HotpotQA with Qwen3.5-4B.
(a) We compare process rewards, outcome rewards, and their combination, with ReAct as a reference.
Process rewards in (a) use verification against gold evidence.
(b) Each verification mechanism adds process rewards
to the outcome-only GRPO baseline.
}
\label{tab:ablations}
\end{table}

\textbf{Effect of process rewards.}
To examine the interaction between outcome and process rewards, we first considered outcome-only GRPO, which improved performance from the ReAct of $46.78$ to $57.73$.
We then used process rewards based on gold-evidence matching to isolate the effect of process rewards.
In Table~\ref{tab:ablations} (a), process-only training achieved $47.27$, suggesting that only process rewards were insufficient to optimize the final task objective.
In contrast, process and outcome rewards further improved performance to $61.40$.
Since outcome rewards directly supervise final-answer correctness, whereas process rewards provide feedback on the intermediate evidence produced along the trajectory.
These results suggest that evidence-based process supervision is most effective as a complementary signal rather than a replacement for the final task objective.
We next compared different sources of process rewards, including our verifier that does not rely on gold evidence.

\textbf{Comparison of Verification Mechanisms.}
We compare three process verification mechanisms.
As shown in Table~\ref{tab:ablations}, our verifier achieves an accuracy of 60.18, outperforming LLM-as-a-judge.
Gold-evidence matching achieves the accuracy at 61.40.
These results suggest that deterministic certificate verification can provide useful process rewards without gold evidence or an LLM judge.

\subsection{Effectiveness Across Policy Optimization Algorithms}
\label{subsec:algorithm}
Finally, we evaluated our plug-and-play verifier framework with both GRPO and PPO on HotpotQA using Qwen3.5-4B to examine whether the benefits of process verification extended beyond GRPO.
For each algorithm, we compared outcome-only training with training augmented by process rewards.
We used the same verifier and reward schedule across the two algorithms, while keeping initialization, task interface, training budget, and evaluation settings fixed.

As shown in Table~\ref{tab:optimization_algorithms},  we observed that incorporating process verification improved accuracy from $57.73$ to $60.18$ under GRPO and from $55.10$ to $57.60$ under PPO.
We obtained these gains using the same verifier module for both algorithms.
Our results indicate that our plug-and-play verifier decoupling process verification from policy optimization enabled the verifier to be reused across GRPO and PPO.

\begin{table}[!htbp]
\centering
\small
\begin{tabular}{@{}lcc@{}}
\toprule
\textbf{Algorithm}
& \textbf{Outcome-only}
& \textbf{+ Verification} \\
\midrule
GRPO & 57.73 & \textbf{60.18} (+2.45) \\
PPO  & 55.10 & \textbf{57.60} (+2.50) \\
\bottomrule
\end{tabular}
\caption{
Effect of process verification under GRPO and PPO
on HotpotQA with Qwen3.5-4B.
We report outcome accuracy (\%).
``+ Verification'' adds process rewards
to outcome-only training.
Values in parentheses denote absolute accuracy gains
over outcome-only training within each algorithm,
in percentage points.
}
\label{tab:optimization_algorithms}
\end{table}

\section{Conclusion}
We propose a plug-and-play verifier framework that enables process-supervised RL for language model agents by decoupling process verification from reward composition and policy optimization.
We note that our framework shares a similar idea with recent works.
\citet{yuan2026verifiable} convert symbolic and algorithmic oracles into turn-level process rewards.
\citet{xu2026stamp} attribute verified evidence to the search actions that first exposed it.
However, our framework is unique in combining a unified contract of bounded weighted sums of verification scores assigned to source steps, deterministic verifiers that require neither gold evidence nor LLM judges, and group-shared reward weight.
Our results illustrate the potential of explicit process verification as a modular complement to outcome rewards in agentic RL.

\textbf{Limitations and future work.}  We evaluate our proposed verifiers by mostly checking per-turn reasoning steps, such as evidence grounding and execution replay, rather than the multiple-turn semantic correctness of intermediate steps.
We will incorporate process reward models~\citep{lightman2024let,setlur2025rewarding} and the most recent calibrated decision models such as Jev for future expansion of the plug-and-play verifier.
Second, our agents operate within short interaction horizons on public agentic benchmarks.
We will extend our framework to much longer-horizon environments~\citep{zhou2024webarena,jimenez2024swe} for production agentic post-training.
Beyond text and image reasoning, we will explore extending our plug-and-play verifier framework to multimodal video generation models~\citep{cui2026joint}.
Finally, we will explore verification-guided RSI, where agents iteratively refine their reasoning strategies and tool-use workflows.

\section*{Disclose Use of LLMs}
We used LLMs, such as Claude and GPT, for initial idea brainstorming and later to check grammar and spelling and improve the clarity of our manuscript's language.
All scientific contents, including figures, tables, experimental design, analysis, and interpretations, are entirely the original work.

\section*{Ethics Statement}
Our research does not involve any potential conflicts of interest or sponsorship issues.
We have carefully considered and addressed issues related to discrimination, bias, and fairness in our methodology.
This study does not involve any privacy or security concerns, fully complies with legal requirements, and upholds the highest standards of research integrity.
All experimental procedures and data processing follow established ethical guidelines for machine learning research.

\section*{Reproducibility Statement}
To support reproducibility, we describe the proposed plug-and-play verification framework and decoupled reward composition.
The experimental settings and evaluation protocol are presented in Section~\ref{subsec:setup}.
Dataset specifications are provided in Appendix~\ref{app:training}, reward construction and policy optimization in Appendix~\ref{app:policy_optimization}, training hyperparameters in Appendix~\ref{app:training_details}, and representative verification inputs and checks in Appendix~\ref{app:example_responses}.
Code is available at \url{https://github.com/momo1443/plug-and-play-verification}.

\bibliography{iclr2027_conference}
\bibliographystyle{iclr2027_conference}

\appendix
\raggedbottom
\section{Training and Evaluation Benchmarks}
\label{app:training}
We conduct training and evaluation using benchmarks from four domains: multi-hop question answering, code generation, multimodal reasoning, and mathematical reasoning.
\renewcommand{\labelitemi}{$\bullet$}
\begin{itemize}
\item{HotpotQA}~\citep{yang2018hotpotqa}: a large-scale multi-hop question answering dataset constructed from Wikipedia.
The dataset contains 90,447 training and 7,405 validation examples, with questions requiring reasoning over information distributed across multiple documents.
Each example is additionally annotated with sentence-level supporting facts, enabling explicit evaluation of evidence identification and multi-hop reasoning.
We use the validation examples of HotpotQA for evaluation to measure the generalization of multi-hop question answering.
\item{TACO}~\citep{li2023taco}: a corpus consisting of 25,443 competitive programming problems from the training split of the TACO dataset.
Each problem is paired with a natural-language problem statement, input-output specifications, reference solutions, and executable test cases, spanning algorithmic topics and difficulty levels for code generation and reasoning.
\item{LiveCodeBench}~\citep{jain2025livecodebench}: a 1,055-problem code generation benchmark consisting of competitive programming problems collected from LeetCode, AtCoder, and Codeforces.
We use the \texttt{code\_generation\_lite} setting with \texttt{release\_v6}, covering problems released between May 2023 and April 2025.
Its temporally recent problems and hidden test cases make it suitable for evaluating coding ability.
We use LiveCodeBench for evaluation to measure the generalization of code generation.
\item{Vision-R1-RL}~\citep{huang2026vision}: a multimodal mathematical reasoning dataset containing approximately 11.3K image-text problems designed for reinforcement learning with verifiable rewards.
The dataset consists primarily of visually grounded mathematical and geometric reasoning problems with deterministic final answers.
\item{MATH-Vision}~\citep{wang2024measuring}: a multimodal mathematical reasoning benchmark consisting of 3,040 visual mathematics problems collected from real-world mathematical competitions.
The benchmark spans 16 mathematical disciplines and 5 difficulty levels, covering diverse problems that require both visual understanding and reasoning.
We use MATH-Vision for evaluation to measure the generalization of multimodal reasoning.
\item{DeepScaleR-Preview-Dataset}~\citep{luo2025deepscaler}:  a curated training corpus of approximately 40,000 unique mathematics problem--answer pairs compiled from AIME problems (1984--2023), AMC problems (before 2023), the Omni-MATH dataset, and the Still dataset, spanning difficulty levels from high-school competition to advanced mathematical reasoning.
\item{AIME 2026}~\citep{dekoninck2026matharena}:  a benchmark comprising 30 problems from the 2026 AIME I and AIME II examinations. 
We use this recent competition benchmark to assess the generalization of mathematical reasoning to challenging competition-level problems.
\end{itemize}

The downloading link for each dataset can be accessed from the Hugging Face website:
\begin{itemize}
\item{HotpotQA}: \url{https://huggingface.co/datasets/hotpotqa/hotpot_qa}
\item{TACO}: \url{https://huggingface.co/datasets/BAAI/TACO}
\item{LiveCodeBench}: \url{https://huggingface.co/datasets/livecodebench/code_generation_lite}
\item{Vision-R1-RL}: \url{https://huggingface.co/datasets/Osilly/Vision-R1-rl}
\item{MATH-Vision}: \url{https://huggingface.co/datasets/MathLLMs/MathVision}
\item{DeepScaleR-Preview-Dataset}: \url{https://huggingface.co/datasets/agentica-org/DeepScaleR-Preview-Dataset}
\item{AIME 2026}: \url{https://huggingface.co/datasets/MathArena/aime_2026}
\end{itemize}

\section{Policy Optimization Details}
\label{app:policy_optimization}
We describe step-level reward construction and the advantage estimators used by GRPO and PPO.
Both algorithms share the same actor objective and policy-token mask, but use different advantage estimators.

\paragraph{Step-level reward construction.}
Consider a training update $k$ and prompt group $g$ for input $x$.
The rollout policy $\pi_{\mathrm{old}}$ generates $G$ trajectories $\{\tau_i\}_{i=1}^{G}$, where $i$ indexes the trajectories within the prompt group.
Let $T_i$ denote the number of interaction steps in $\tau_i$, $E_i=E_d(x,\tau_i)$ its outcome reward, and $p_{i,t}=p_{d,t}(x,\tau_i)$ its step-level process reward.
Following the reward composition in Eq.~\eqref{eq:reward_composition}, we define
\begin{equation}
r_{i,t}=
\begin{cases}
\mathbf{1}\{t=T_i\}E_i,
& k\leq K_{\mathrm{w}},\\[3pt]
w_{k,g}\mathbf{1}\{t=T_i\}E_i+(1-w_{k,g})p_{i,t},
& k>K_{\mathrm{w}}.
\end{cases}
\label{eq:step_mixed_reward}
\end{equation}
Here $\mathbf{1}\{\cdot\}$ denotes the indicator function, $K_{\mathrm{w}}=50$ is the number of outcome-only warmup updates, and $w_{k,g}\sim\mathcal U(0,1)$ is a reproducibly sampled weighted coefficient shared by all rollouts within prompt group $g$.
We omit the fixed update $k$ and group $g$ in $r_{i,t}$ for readability.
The outcome reward is assigned only to the terminal interaction step, while process rewards are assigned to their source interaction steps.
Because $\sum_t p_{i,t}=P_d(x,\tau_i)$, summing these composed step-level rewards yields the trajectory-level reward $R_{k,i}$.
For incomplete trajectories, $E_i=0$, while independently verified process rewards are retained.

\paragraph{Group-relative advantages.}
For GRPO, we use undiscounted returns:
\begin{equation}
R_{i,t}=\sum_{u=t}^{T_i}r_{i,u}.
\label{eq:reward_to_go}
\end{equation}
Each reward contributes to the returns of its assigned step and all preceding steps, regardless of when verification is performed.
The terminal outcome reward is therefore included in each valid step's return, although it is assigned only once to the terminal step.

We normalize returns across trajectories from the same prompt that reach interaction step $t$:
\begin{equation}
\hat A_{i,t}
=
\frac{R_{i,t}-\bar R_t}{s_t+\epsilon_{\mathrm{adv}}},
\label{eq:group_advantage}
\end{equation}
where $\bar R_t$ and $s_t$ are the mean and sample standard deviation of the corresponding returns, respectively.
For a group containing only one trajectory at step $t$, we set $\bar R_t=0$ and $s_t=1$ to retain its return.
Outcome and process rewards are combined before computing returns and advantages, rather than normalized separately.
The resulting advantages remain fixed during policy optimization.

\paragraph{Last-mile signal under outcome rewards.}
Consider trajectories within the same prompt group that share outcome reward $E$ and reward weight $w<1$.
Let $S_{i,t}=\sum_{u=t}^{T_i}p_{i,u}$ denote the remaining process return.
The composed return is $R_{i,t}=wE+(1-w)S_{i,t}$.
For groups with more than one trajectory, substituting this into the group-relative advantage gives
\begin{equation}
\hat A_{i,t}
=
\frac{S_{i,t}-\bar S_t}
{s_{S,t}+\epsilon_{\mathrm{adv}}/(1-w)},
\end{equation}
where $\bar S_t$ and $s_{S,t}$ are the mean and sample standard deviation of the process returns in the group.
This recovers Eq.~\eqref{eq:tied_outcome_advantage}.
The shared outcome reward cancels under group centering, allowing differences in verified process returns to distinguish trajectories with identical outcomes.
When $w=1$, identical outcome rewards instead produce zero group-relative advantages.

\paragraph{Policy-token masking and actor optimization.}
We assign each step-level advantage to all policy-generated tokens within that step, including reasoning text, tool-call arguments, and final-answer tokens.
Prompt tokens and environment observations are excluded from the actor and KL losses, while observations remain available as context for subsequent policy generations.

For a policy token $z$ with context $h$, we define the importance ratio
\begin{equation}
\rho
=
\frac{\pi_\theta(z\mid h)}
{\pi_{\mathrm{old}}(z\mid h)}.
\label{eq:policy_ratio}
\end{equation}
Using its assigned step advantage $A$, the clipped surrogate is
\begin{equation}
\ell_{\mathrm{clip}}
=
\min\left(
\rho A,\,
\operatorname{clip}(\rho,1-\epsilon,1+\epsilon)A
\right).
\label{eq:clipped_surrogate}
\end{equation}
We additionally apply dual clipping with $c=3$: $\ell=\max(\ell_{\mathrm{clip}},3A)$ when $A<0$, and $\ell=\ell_{\mathrm{clip}}$ otherwise.

Let $\langle\cdot\rangle$ denote averaging first over valid policy tokens within each interaction step and then over all valid interaction steps in the sampled prompt group.
We maximize
\begin{equation}
\widehat{\mathcal J}(\theta)
=
\langle\ell\rangle
-\beta\langle e^\Delta-\Delta-1\rangle,
\label{eq:policy_objective}
\end{equation}
where $\Delta=\log\pi_{\mathrm{ref}}(z\mid h) -\log\pi_\theta(z\mid h)$, and $\beta$ controls KL regularization.
The reference policy remains fixed during training.
KL regularization is applied only to the actor objective, not to advantage estimation.

\paragraph{Integration with PPO.}
PPO uses the same verifier outputs, composed step rewards, policy-token mask, and clipped actor objective as GRPO.
The difference lies in advantage estimation: PPO learns a critic and uses generalized advantage estimation (GAE) instead of group-relative normalization.

With $\gamma=\lambda=1$ and zero terminal bootstrapping, step-level GAE simplifies to
\begin{equation}
A^{\mathrm{GAE}}_{i,t}
=
R_{i,t}-V_{\phi_{\mathrm{old}}}(s_{i,t}),
\label{eq:ppo_step_gae}
\end{equation}
where $s_{i,t}$ denotes the interaction history before policy generation at step $t$ in trajectory $\tau_i$, and $V_{\phi_{\mathrm{old}}}$ is the value function parameterized by the pre-update critic parameters $\phi_{\mathrm{old}}$.
We standardize these advantages across all valid interaction steps in the rollout batch:
\begin{equation}
\hat A^{\mathrm{PPO}}_{i,t}
=
\frac{A^{\mathrm{GAE}}_{i,t}-\bar A}
{s_A+\epsilon_{\mathrm{adv}}},
\label{eq:ppo_advantage_normalization}
\end{equation}
where $\bar A$ and $s_A$ denote the batch mean and sample standard deviation, respectively.
These normalized advantages replace GRPO's group-relative advantages in the shared actor objective.

The critic target is the undiscounted return:
\begin{equation}
Y_{i,t}=R_{i,t}.
\label{eq:ppo_value_target}
\end{equation}
We train the critic using a clipped squared-error value loss with clipping range $0.5$, evaluated once per interaction-step starting state.
Actor and critic parameters are updated separately, without a critic-only warmup.

In the Qwen3.5-4B HotpotQA comparison, the actor and critic learning rates are $10^{-6}$ and $10^{-5}$, respectively.

\section{Training Details}
\label{app:training_details}

\paragraph{Training infrastructure.}
All experiments are conducted on a single Kubernetes (K8s) node with $8\times$ NVIDIA A40 GPUs (48 GB each).
We use PyTorch with Fully Sharded Data Parallelism (FSDP) for distributed training and vLLM in server mode for rollout generation.
A run of $500$ training updates takes approximately $80$ hours for Qwen3.5-4B and $94$ hours for Qwen3.5-9B.

\paragraph{Training data and rollout budgets.}
We select the first $10{,}000$ examples from each prepared training split and train for $500$ updates per domain.
Both GRPO configurations sample four trajectories per prompt.
The global prompt batch size and actor mini-batch size are $20$ prompts for Qwen3.5-4B and $8$ prompts for Qwen3.5-9B.
Consequently, a $500$-update run samples $10{,}000$ and $4{,}000$ prompt instances, respectively, before expanding each prompt into four trajectories.

\begin{table}[H]
\centering
\small
\captionsetup{justification=centering}
\begin{minipage}[t]{0.49\linewidth}
\centering
\scalebox{0.9}{%
\begin{tabular}{@{}ll@{}}
\toprule
\textbf{Hyperparameter} & \textbf{Value} \\
\midrule
Optimizer & AdamW \\
Learning rate & $10^{-6}$ \\
Training updates & 500 \\
LR schedule & Constant; no warmup \\
Global prompt batch size & 20 \\
Trajectories per prompt & 4 \\
Mini-batch size & 20 prompts \\
Optimization epochs per rollout & 1 \\
Initial prompt limit & 2048 tokens \\
History / context limit & 8192 tokens \\
Maximum interaction steps & 5 \\
Sampling temperature & 1.0 \\
Top-$p$ / Top-$k$ & 1.0 / -1 \\
Weight decay & 0.01 \\
Gradient clipping & 1.0 \\
KL coefficient & 0.001 \\
Precision & bfloat16 \\
Random seed & 42 \\
\bottomrule
\end{tabular}%
}
\caption{Hyperparameters for Qwen3.5-4B GRPO training on DeepScaleR.}
\label{tab:qwen35_4b}
\end{minipage}\hfill
\begin{minipage}[t]{0.49\linewidth}
\centering
\scalebox{0.9}{%
\begin{tabular}{@{}ll@{}}
\toprule
\textbf{Hyperparameter} & \textbf{Value} \\
\midrule
Fine-tuning method & LoRA ($r=64$, $\alpha=64$) \\
Optimizer & AdamW \\
Learning rate & $10^{-6}$ \\
Training updates & 500 \\
LR schedule & Constant; no warmup \\
Global prompt batch size & 8 \\
Trajectories per prompt & 4 \\
Mini-batch size & 8 prompts \\
Optimization epochs per rollout & 1 \\
Initial prompt limit & 2048 tokens \\
History / context limit & 8192 tokens \\
Maximum interaction steps & 5 \\
Sampling temperature & 1.0 \\
Top-$p$ / Top-$k$ & 1.0 / -1 \\
Weight decay & 0.01 \\
Gradient clipping & 1.0 \\
KL coefficient & 0.001 \\
Precision & bfloat16 \\
LoRA target modules
& \texttt{q\_proj}, \texttt{k\_proj}, \\
& \texttt{v\_proj}, \texttt{o\_proj}, \\
& \texttt{up\_proj}, \texttt{gate\_proj}, \\
& \texttt{down\_proj} \\
Random seed & 42 \\
\bottomrule
\end{tabular}%
}
\caption{Hyperparameters for Qwen3.5-9B LoRA GRPO training on DeepScaleR.}
\label{tab:qwen35_9b_grpo_hyperparams}
\end{minipage}
\end{table}

\begin{table}[H]
\centering
\small
\captionsetup{justification=centering}
\begin{minipage}[t]{0.49\linewidth}
\centering
\scalebox{0.9}{%
\begin{tabular}{@{}ll@{}}
\toprule
\textbf{Hyperparameter} & \textbf{Value} \\
\midrule
Optimizer & AdamW \\
Learning rate & $10^{-6}$ \\
Training updates & 500 \\
LR schedule & Constant; no warmup \\
Global prompt batch size & 20 \\
Trajectories per prompt & 4 \\
Mini-batch size & 20 prompts \\
Optimization epochs per rollout & 1 \\
Max prompt length & 8192 tokens \\
Max completion length & 1024 tokens \\
Maximum interaction steps & 4 \\
Sampling temperature & 1.0 \\
Top-$p$ / Top-$k$ & 1.0 / -1 \\
Weight decay & 0.01 \\
Gradient clipping & 1.0 \\
KL coefficient & 0.001 \\
Precision & bfloat16 \\
Random seed & 42 \\
\bottomrule
\end{tabular}%
}
\caption{Hyperparameters for Qwen3.5-4B GRPO training on HotpotQA.}
\label{tab:qwen35_4b_hotpotqa}
\end{minipage}\hfill
\begin{minipage}[t]{0.49\linewidth}
\centering
\scalebox{0.9}{%
\begin{tabular}{@{}ll@{}}
\toprule
\textbf{Hyperparameter} & \textbf{Value} \\
\midrule
Fine-tuning method & LoRA ($r=64$, $\alpha=64$) \\
Optimizer & AdamW \\
Learning rate & $10^{-6}$ \\
Training updates & 500 \\
LR schedule & Constant; no warmup \\
Global prompt batch size & 8 \\
Trajectories per prompt & 4 \\
Mini-batch size & 8 prompts \\
Optimization epochs per rollout & 1 \\
Max prompt length & 8192 tokens \\
Max completion length & 1024 tokens \\
Maximum interaction steps & 4 \\
Sampling temperature & 1.0 \\
Top-$p$ / Top-$k$ & 1.0 / -1 \\
Weight decay & 0.01 \\
Gradient clipping & 1.0 \\
KL coefficient & 0.001 \\
Precision & bfloat16 \\
LoRA target modules
& \texttt{q\_proj}, \texttt{k\_proj}, \\
& \texttt{v\_proj}, \texttt{o\_proj}, \\
& \texttt{up\_proj}, \texttt{gate\_proj}, \\
& \texttt{down\_proj} \\
Random seed & 42 \\
\bottomrule
\end{tabular}%
}
\caption{Hyperparameters for Qwen3.5-9B LoRA GRPO training on HotpotQA.}
\label{tab:qwen35_9b_hotpotqa}
\end{minipage}
\end{table}

\begin{table}[H]
\centering
\small
\captionsetup{justification=centering}
\begin{minipage}[t]{0.49\linewidth}
\centering
\scalebox{0.9}{%
\begin{tabular}{@{}ll@{}}
\toprule
\textbf{Hyperparameter} & \textbf{Value} \\
\midrule
Optimizer & AdamW \\
Learning rate & $10^{-6}$ \\
Training updates & 500 \\
LR schedule & Constant; no warmup \\
Global prompt batch size & 20 \\
Trajectories per prompt & 4 \\
Mini-batch size & 20 prompts \\
Optimization epochs per rollout & 1 \\
Max prompt length & 8192 tokens \\
Max completion length & 2048 tokens \\
Maximum interaction steps & 5 \\
Sampling temperature & 1.0 \\
Top-$p$ / Top-$k$ & 1.0 / -1 \\
Weight decay & 0.01 \\
Gradient clipping & 1.0 \\
KL coefficient & 0.001 \\
Precision & bfloat16 \\
Random seed & 42 \\
\bottomrule
\end{tabular}%
}
\caption{Hyperparameters for Qwen3.5-4B GRPO training on TACO.}
\label{tab:qwen35_4b_taco}
\end{minipage}\hfill
\begin{minipage}[t]{0.49\linewidth}
\centering
\scalebox{0.9}{%
\begin{tabular}{@{}ll@{}}
\toprule
\textbf{Hyperparameter} & \textbf{Value} \\
\midrule
Fine-tuning method & LoRA ($r=64$, $\alpha=64$) \\
Optimizer & AdamW \\
Learning rate & $10^{-6}$ \\
Training updates & 500 \\
LR schedule & Constant; no warmup \\
Global prompt batch size & 8 \\
Trajectories per prompt & 4 \\
Mini-batch size & 8 prompts \\
Optimization epochs per rollout & 1 \\
Max prompt length & 8192 tokens \\
Max completion length & 2048 tokens \\
Maximum interaction steps & 5 \\
Sampling temperature & 1.0 \\
Top-$p$ / Top-$k$ & 1.0 / -1 \\
Weight decay & 0.01 \\
Gradient clipping & 1.0 \\
KL coefficient & 0.001 \\
Precision & bfloat16 \\
LoRA target modules
& \texttt{q\_proj}, \texttt{k\_proj}, \\
& \texttt{v\_proj}, \texttt{o\_proj}, \\
& \texttt{up\_proj}, \texttt{gate\_proj}, \\
& \texttt{down\_proj} \\
Random seed & 42 \\
\bottomrule
\end{tabular}%
}
\caption{Hyperparameters for Qwen3.5-9B LoRA GRPO training on TACO.}
\label{tab:qwen35_9b_taco}
\end{minipage}
\end{table}

\begin{table}[H]
\centering
\small
\captionsetup{justification=centering}
\begin{minipage}[t]{0.49\linewidth}
\centering
\scalebox{0.9}{%
\begin{tabular}{@{}ll@{}}
\toprule
\textbf{Hyperparameter} & \textbf{Value} \\
\midrule
Optimizer & AdamW \\
Learning rate & $10^{-6}$ \\
Training updates & 500 \\
LR schedule & Constant; no warmup \\
Global prompt batch size & 20 \\
Trajectories per prompt & 4 \\
Mini-batch size & 20 prompts \\
Optimization epochs per rollout & 1 \\
Max prompt length & 8192 tokens \\
Max completion length & 2048 tokens \\
Maximum interaction steps & 3 \\
Sampling temperature & 1.0 \\
Top-$p$ / Top-$k$ & 1.0 / -1 \\
Weight decay & 0.01 \\
Gradient clipping & 1.0 \\
KL coefficient & 0.001 \\
Precision & bfloat16 \\
Random seed & 42 \\
\bottomrule
\end{tabular}%
}
\caption{Hyperparameters for Qwen3.5-4B GRPO training on Vision-R1.}
\label{tab:qwen35_4b_vision}
\end{minipage}\hfill
\begin{minipage}[t]{0.49\linewidth}
\centering
\scalebox{0.9}{%
\begin{tabular}{@{}ll@{}}
\toprule
\textbf{Hyperparameter} & \textbf{Value} \\
\midrule
Fine-tuning method & LoRA ($r=64$, $\alpha=64$) \\
Optimizer & AdamW \\
Learning rate & $10^{-6}$ \\
Training updates & 500 \\
LR schedule & Constant; no warmup \\
Global prompt batch size & 8 \\
Trajectories per prompt & 4 \\
Mini-batch size & 8 prompts \\
Optimization epochs per rollout & 1 \\
Max prompt length & 8192 tokens \\
Max completion length & 2048 tokens \\
Maximum interaction steps & 3 \\
Sampling temperature & 1.0 \\
Top-$p$ / Top-$k$ & 1.0 / -1 \\
Weight decay & 0.01 \\
Gradient clipping & 1.0 \\
KL coefficient & 0.001 \\
Precision & bfloat16 \\
LoRA target modules
& \texttt{q\_proj}, \texttt{k\_proj}, \\
& \texttt{v\_proj}, \texttt{o\_proj}, \\
& \texttt{up\_proj}, \texttt{gate\_proj}, \\
& \texttt{down\_proj} \\
Random seed & 42 \\
\bottomrule
\end{tabular}%
}
\caption{Hyperparameters for Qwen3.5-9B LoRA GRPO training on Vision-R1.}
\label{tab:qwen35_9b_vision}
\end{minipage}
\end{table}

We use the same experimental setup, verifier, and reward composition for PPO and GRPO.
PPO instead uses step-level GAE and a separately optimized critic initialized from the same pretrained checkpoint as the actor.
The outcome-only PPO baseline fixes $w=1$, whereas verification PPO follows the same group-shared reward weight schedule as verification GRPO.
Validation uses outcome-only rewards, and no critic-only warmup is performed.
Both PPO variants use the hyperparameters in Table~\ref{tab:qwen35_4b_hotpotqa_ppo}.

\begin{table}[H]
\centering
\small
\begin{tabular}{@{}p{0.58\linewidth}p{0.35\linewidth}@{}}
\toprule
\textbf{Hyperparameter} & \textbf{Value} \\
\midrule
Actor / critic optimizer & AdamW \\
Actor learning rate & $10^{-6}$ \\
Critic learning rate & $10^{-5}$ \\
Actor / critic LR schedule & Constant; no LR warmup \\
Training updates & 500 \\
Global prompt batch size & 20 \\
Trajectories per prompt & 4 \\
Max prompt length & 8192 tokens \\
Max completion length & 1024 tokens \\
Top-$p$ / Top-$k$ & 1.0 / $-1$ \\
Discount factor $\gamma$ / GAE parameter $\lambda$ & 1.0 / 1.0 \\
Policy clipping threshold & 0.2 \\
Dual-clip constant & 3.0 \\
Value clipping range & 0.5 \\
Actor / critic weight decay & 0.01 \\
Actor / critic gradient clipping & 1.0 \\
Actor-loss KL coefficient & 0.001 \\
Random seed & 42 \\
\bottomrule
\end{tabular}
\caption{Hyperparameters for Qwen3.5-4B PPO training on HotpotQA.}
\label{tab:qwen35_4b_hotpotqa_ppo}
\end{table}

\begin{table}[htbp]
\centering
\begin{tabular}{@{}ll@{}}
\toprule
\textbf{Parameter} & \textbf{Value} \\
\midrule
Sampling temperature & 0.6 \\
Top-$p$ / Top-$k$ & 1.0 / -1 \\
Precision / Random seed & bfloat16 / 42 \\
Inference engine & vLLM \\
\bottomrule
\end{tabular}
\caption{Evaluation settings shared by all benchmarks.}
\label{tab:eval_config}
\end{table}

\begin{table*}[htbp]
\centering

\begin{tabular}{@{}lllrrrrr@{}}
\toprule
\textbf{Benchmark}
& \textbf{Metric}
& \shortstack{\textbf{Thinking}\\\textbf{mode}}
& \textbf{Steps}
& \shortstack{\textbf{Per-step}\\\textbf{cap}}
& \shortstack{\textbf{Total}\\\textbf{budget}}
& \shortstack{\textbf{Prompt}\\\textbf{limit}}
& \textbf{Context} \\
\midrule
AIME 2025
& avg@1 & On & 5
& 2{,}048 & 10{,}240 & 2{,}048 & 13{,}312 \\

AIME 2026
& avg@4 & On & 5
& 2{,}048 & 10{,}240 & 2{,}048 & 13{,}312 \\

HotpotQA
& avg@1 & Off & 4
& 1{,}024 & 4{,}096 & 8{,}192 & 12{,}288 \\

LiveCodeBench v6
& avg@4 & Off & 5
& 2{,}048 & 10{,}240 & 8{,}192 & 12{,}288 \\

MATH-Vision
& avg@1 & Off & 3
& 2{,}048 & 6{,}144 & 8{,}192 & 12{,}288 \\
\bottomrule
\end{tabular}

\vspace{2pt}
\caption{Evaluation protocol across all benchmarks.
avg@$k$ denotes accuracy averaged over $k$ independently sampled
trajectories per problem.
The same sampling protocol is applied to all compared methods.}
\label{tab:eval_budget}
\end{table*}

\section{Representative Verification Inputs and Checks}
\label{app:example_responses}
We present representative examples across four reasoning domains to illustrate how intermediate artifacts are verified independently of final-answer correctness.

\subsection{Multi-hop Question Answering: HotpotQA}
\label{app:example_qa}
\begin{questionbox}
\textbf{Question:}
What 2014 American 3D computer-animated musical fantasy adventure comedy film features a song recorded by British singer Leah McFall?

\textbf{Reference answer:} The Book of Life.
\end{questionbox}
\begin{responsebox}
\textbf{Illustrative Agent Trajectory.}
The following trajectory is constructed from the recorded question and retrieved evidence to illustrate the local verification process.

\noindent\textbf{Step 1: Initial Search.}
The agent searches for ``2014 American 3D computer-animated musical fantasy adventure comedy film song recorded by British singer Leah McFall''.
The retrieved results include \texttt{passage:9904} (\emph{The Book of Life (2014 film)}) and \texttt{passage:9898} (\emph{Home (Leah McFall song)}).
The latter contains the supporting sentence:
\begin{quote}
``This song also featured in the latest film, \texttt{"The Book of Life"}.''
\end{quote}
We denote this sentence by $S$.
As the initial search requires no certificate, this step is excluded from verification and normalization.

\noindent\textbf{Step 2: Evidence-Grounded Search.}
The agent searches for ``The Book of Life Leah McFall'' and provides a certificate containing:
\begin{itemize}
    \item \textbf{Source:} \texttt{passage:9898};
    \item \textbf{Evidence:} Sentence $S$;
    \item \textbf{Target:} \texttt{"The Book of Life"}.
\end{itemize}

\noindent\textbf{Step 3: Final Answer.}
The agent submits \texttt{"The Book of Life"} with a final certificate citing \texttt{passage:9898}, sentence $S$, and the answer span \texttt{"The Book of Life"}.

\noindent\textbf{Verification Results.}
At Step 2, source-span grounding succeeds because $S$ appears in the cited passage, while action-certificate matching succeeds because the certified target is consistent with the search action.
At Step 3, both checks also pass: the cited evidence is grounded in the source, and the certified answer span matches the submitted answer.Therefore, all applicable local checks pass, yielding $C_d(x,\tau)=1$ under Eq.~\eqref{eq:verified_self_consistency}.
\end{responsebox}

\subsection{Code Generation: LiveCodeBench}
\label{app:example_code}

\begin{questionbox}
\textbf{Question:} Given an integer array, find the maximum sum of a nonempty contiguous segment whose adjacent elements have opposite parity.

\textbf{Example input--output test cases:}
$[1,2,3,4,5]\mapsto15$; $[9,9,8,8]\mapsto17$; $[-1000]\mapsto-1000$.
\end{questionbox}

\begin{responsebox}
This is a constructed illustration, not a logged training trajectory.
The solution uses dynamic programming to maintain the maximum valid subarray sum, extending a segment only when adjacent elements have opposite parity.

\begin{verbatim}
best = current = a[0]
for i in range(1, len(a)):
    if a[i] % 2 != a[i - 1] % 2:
        current = max(a[i], current + a[i])
    else:
        current = a[i]
    best = max(best, current)
\end{verbatim}

\textbf{Output:} \texttt{best}.
The algorithm runs in $O(n)$ time and uses $O(1)$ auxiliary space.

\textbf{Verification inputs and local checks.}
On these three developer tests, the code returns $15$, $17$, and $-1000$, respectively.
All three tests pass, yielding \texttt{passed=3}, \texttt{total=3}, and \texttt{all\_passed=true}.
The recorded execution identifies the immutable code artifact and its digest, the developer-test suite and its digest, and the execution-result digest.
The code verifier replays the recorded execution and checks whether the replayed outputs match the recorded execution results.
The submission certificate identifies the final code artifact and its digest and cites this execution record.
The verifier checks recorded execution lineage and replay consistency independently of final submission, assigning verified process rewards to their source execution steps.
Submission-certificate checks are applied only when a final code artifact is submitted.
\end{responsebox}

\subsection{Multimodal Reasoning: MATH-Vision}
\label{app:example_vision}

\begin{questionbox}
\textbf{Question:} Using the balanced scale, determine how many plums have the same weight as one apple.

\medskip \textbf{Reference answer:} $3$.
\end{questionbox}

\begin{responsebox}
This is a constructed illustration, not a logged training trajectory.
The left pan contains three apples and two plums, and the right pan contains one apple and eight plums.
Let $A$ and $P$ denote the weights of an apple and a plum, respectively.
The balance gives
\[
3A+2P=A+8P,\qquad 2A=6P,\qquad A=3P.
\]
Thus, one apple weighs as much as three plums.

\textbf{Final answer:} $\boxed{3}$.

\textbf{Verification inputs and local checks.}
For the original $765\times225$ image, the left- and right-pan crop boxes are $(35,25,304,131)$ and $(474,28,734,130)$, respectively, in $(\mathrm{left},\mathrm{top},\mathrm{right},\mathrm{bottom})$ pixel coordinates.
The visual certificate cites both crop identifiers and includes the final answer $3$ in its claim.
The verifier checks crop provenance and replay consistency against the original image, together with string-level coupling between the submitted answer and the certificate claim.
\end{responsebox}

\subsection{Mathematical Reasoning: AIME 2026}
\label{app:example_math}

\begin{questionbox}
\textbf{Question:} An integer arithmetic progression starts at $4$ and includes $24$ and $34$. Sum the tenth terms over all progressions satisfying these conditions.

\textbf{Reference answer:} $178$.
\end{questionbox}

\begin{responsebox}
This constructed three-step illustration fits within the five-step limit; it is not a logged training trajectory.

\textbf{Step 1: Constraining the common difference.}
Let $d$ be the common difference.
Since all terms are integers and the progression reaches values above $4$, $d$ is a positive integer.
The required terms give
\[
24-4=20,\qquad 34-4=30.
\]
Hence, $d$ divides both $20$ and $30$.

\textbf{Step 2: Enumerating the possibilities.}
The admissible common differences are $1$, $2$, $5$, and $10$, with
\[
1+2+5+10=18.
\]

\textbf{Step 3: Final submission.}
Each progression has tenth term $4+9d$, so their sum is
\[
4\times4+9\times18=178.
\]
\textbf{Final answer:} $\boxed{178}$.

\textbf{Verification inputs and local checks.}
Excluding the final-answer span, the four numeric-equality artifacts are grouped by source step:
\begin{verbatim}
Step 1: 24 - 4 = 20; 34 - 4 = 30
Step 2: 1 + 2 + 5 + 10 = 18
Step 3: 4 * 4 + 9 * 18 = 178
\end{verbatim}
All four equalities hold, with source-step indices $(1,1,2,3)$.
These checks verify arithmetic only, not the divisibility argument, completeness of enumeration, or tenth-term formula.
\end{responsebox}

\vspace{6pt}

\end{document}